\documentclass[letterpaper, 10 pt, conference]{ieeeconf}  

\IEEEoverridecommandlockouts                              

\usepackage{graphics} 
\usepackage{epsfig} 
\usepackage{mathptmx} 
\usepackage{times} 
\usepackage{amsmath} 
\usepackage{amssymb}  
\usepackage{multicol,multirow}
\usepackage{mathtools}
\usepackage{nccmath}
\usepackage{epsfig}
\usepackage{subcaption}

\title{\LARGE \bf
Floorplan-Aware Camera Poses Refinement
}

\author{Anna Sokolova$^1$, Filipp Nikitin$^1$, Anna Vorontsova$^1$, Anton Konushin$^1$
\thanks {$^1$ All authors are with Samsung AI Center, Moscow, Russia, 
{\tt\small \{a.sokolova, f.nikitin, a.vorontsova, a.konushin\}@samsung.com}
}
}

\begin{document}

\maketitle
\thispagestyle{empty}
\pagestyle{empty}

\begin{abstract}

Processing large indoor scenes is a challenging task, as scan registration and camera trajectory estimation methods accumulate errors across time. As a result, the quality of reconstructed scans is insufficient for some applications, such as visual-based localization and navigation, where the correct position of walls is crucial.

For many indoor scenes, there exists an image of a technical floorplan that contains information about the geometry and main structural elements of the scene, such as walls, partitions, and doors. We argue that such a floorplan is a useful source of spatial information, which can guide a 3D model optimization.

The standard RGB-D 3D reconstruction pipeline consists of a tracking module applied to an RGB-D sequence and a bundle adjustment (BA) module that takes the posed RGB-D sequence and corrects the camera poses to improve consistency. We propose a novel optimization algorithm expanding conventional BA that leverages the prior knowledge about the scene structure in the form of a floorplan. Our experiments on the Redwood dataset and our self-captured data demonstrate that utilizing floorplan improves accuracy of 3D reconstructions.

\end{abstract}

\section{INTRODUCTION}

Restoring general scene structure formed with floor and walls is complicated for multiple reasons. First, both floor and walls are often textureless or covered with repetitive patterns, so the keypoints cannot be detected or correctly matched across different frames. Then, the floor and walls might not superimpose after a loop closure in BA due to the errors accumulated over time. Alternatively, the surfaces might not match perfectly when aligning partial scans of large-scale scenes. Either way, multiple duplicate layers appear, making the overall scan corrupted; we refer to this unwanted effect as to \emph{layering}. In addition, each surface might have hills and pits, worsening the visual impression; we call it \emph{unevenness}. Hence, the reconstructed scans come imperfect and should be additionally optimized.

Overall, no-reference approaches are limited by design, so a significant improvement cannot be achieved without additional information about the scene. We argue that a technical floorplan of a scene is one of the most available, intuitive, and easy-to-use sources of spatial data. Floorplans reflect the general structure of the scene, so we can use them as guidance during optimization, comparing the reconstructed scan with a floorplan and penalizing their divergence. 

\begin{figure}[ht!]
    \centering
    \setlength{\tabcolsep}{1pt}
    \begin{tabular}{cc}
    \includegraphics[width=0.49\linewidth]{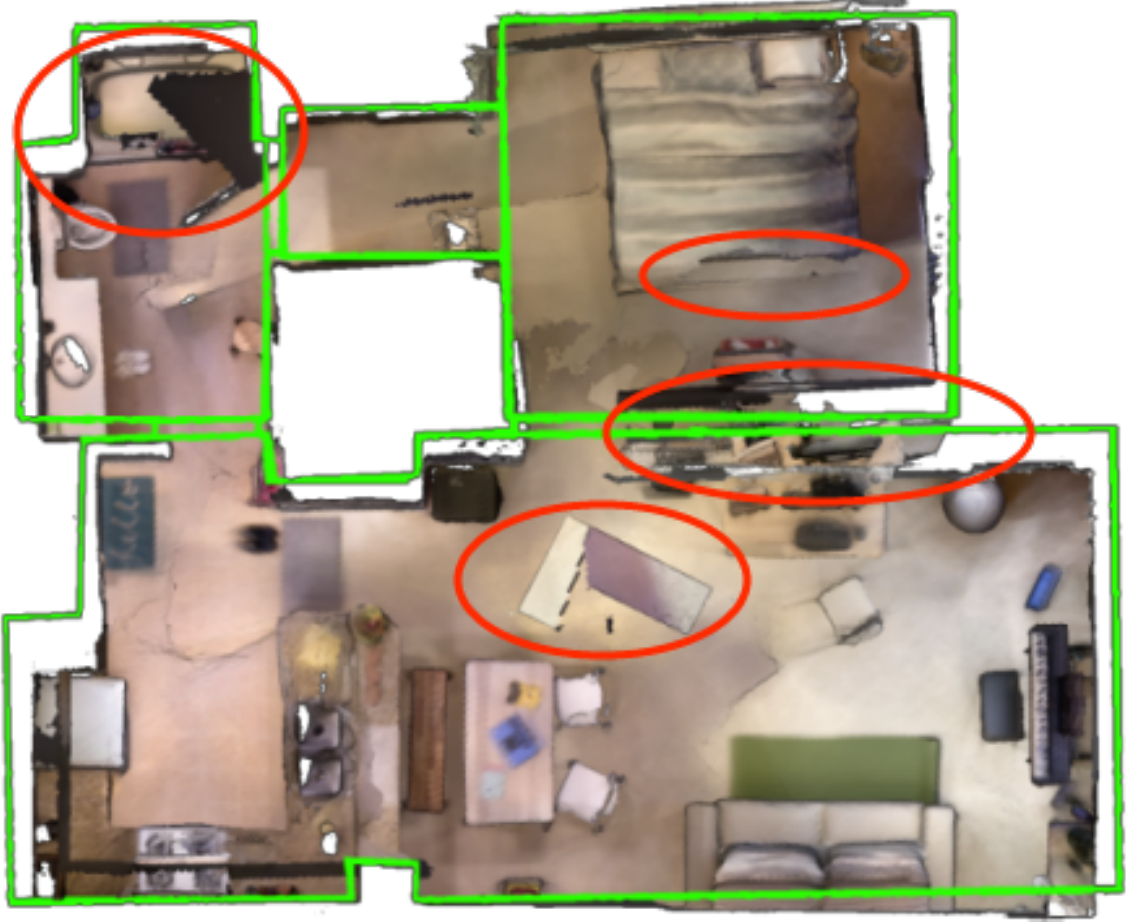} & 
    \includegraphics[width=0.49\linewidth]{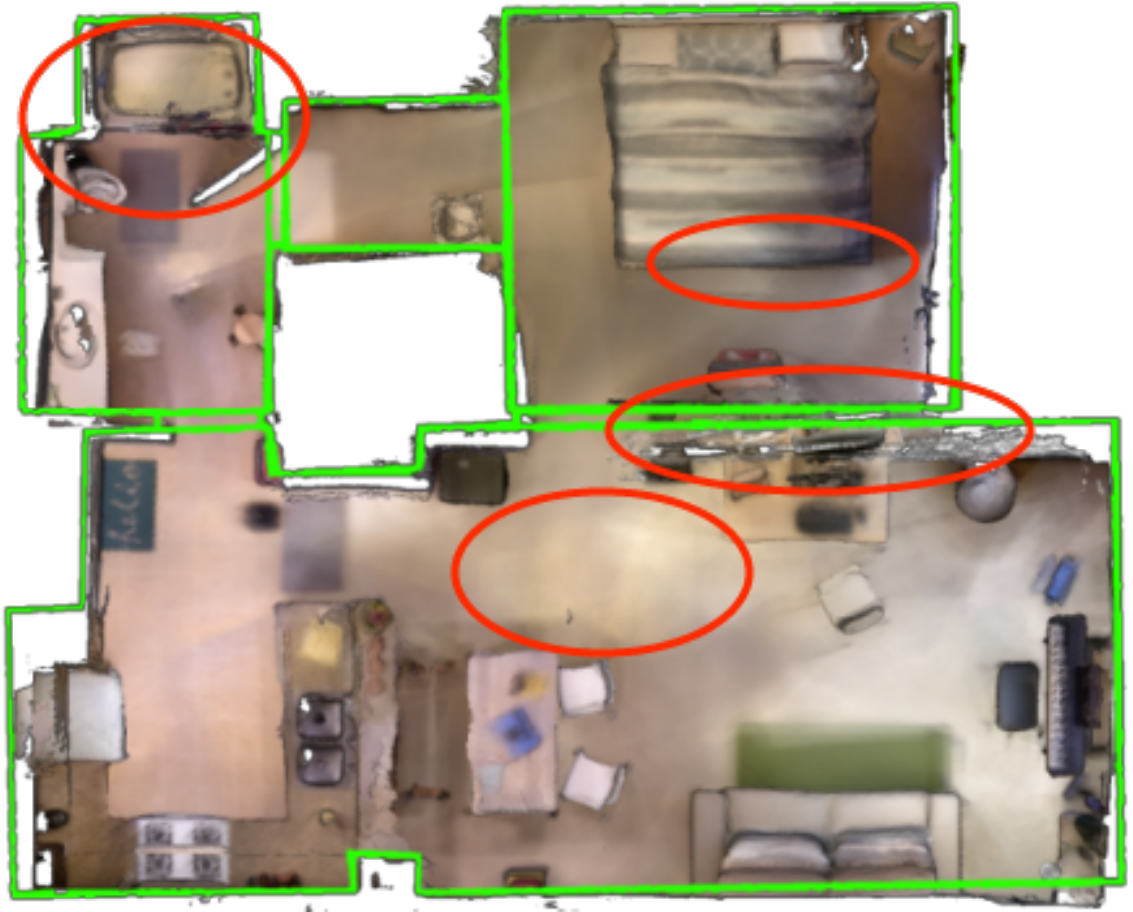}\\
    \end{tabular}
    \caption{The reconstructed scan before (left) and after (right) camera poses refinement with a floorplan guidance. Through refinement, the misplaced upper right room gets aligned with the floorplan, and multiple reconstruction artifacts (marked with red ellipses) decrease or disappear.}
    \label{fig:alignment}
\end{figure}

Accordingly, we address the following problem: given a posed RGB-D sequence and a floorplan, refine camera poses so that the scan reconstructed using these poses is consistent with the floorplan. We assume that we have a floorplan image that depicts vertical architectural surfaces comprising the general scene structure (Fig.~\ref{fig:floorplan_image}). The coordinate transformation (scale, shift, and rotation) between a scan and its floorplan might be unknown.

Typically, in scan reconstruction, camera poses are estimated roughly and then refined using a bundle adjustment (BA). We propose a novel optimization algorithm that expands BA using prior knowledge about the scene structure. We assume that the floor surface is planar, and a scene is bounded with planar walls matching the walls on the floorplan. To obtain a scan that satisfies these requirements, we impose additional constraints in BA. Specifically, we apply semantic segmentation to select points corresponding to floor and walls and penalize floor unevenness and the divergence between the walls and the floorplan.

\section{RELATED WORK}

We propose a floorplan-aware camera poses refinement method which extends BA. We aim to align the scan with the floorplan and also improve geometric consistency. Besides, we rely on semantic segmentation to detect a floor and walls in the scan. Therefore, we review existing formulations of geometric consistency, semantic-based pose refinement, and floorplan-aware 3D reconstruction.

\begin{figure*}[t]
    \centering
    \includegraphics[width=0.975\linewidth]{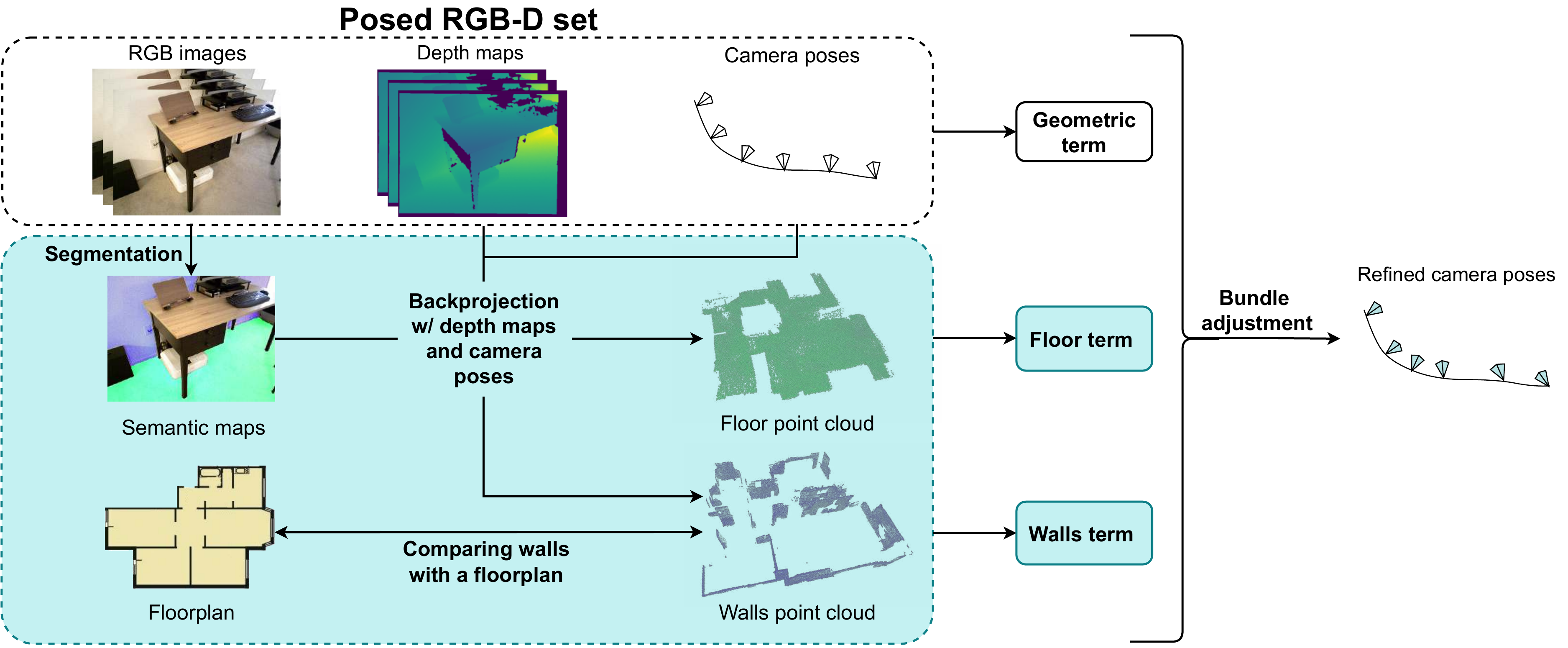}
    \caption{The scheme of the proposed camera poses refinement method. The novel modules and terms are colored turquoise.}
    \label{fig:scheme}
\end{figure*}

\subsection{Geometric Consistency}

The reconstructed scan should be geometry consistent, so scan optimization (known as BA) minimizes the discrepancy between different measurements. The BA term that reflects geometric inconsistency can be formalized in various ways depending on the input data, the model of a scene, and possible applications. One of the most popular geometric terms is based on reprojection error. However, reprojection-based functions are not defined everywhere and exhibit singularities, making the optimization process sensitive to initial conditions and outliers. 
In alternative BA formulations~\cite{hessflores2011ray,schinstock2009alternative}, the cost function is based on the minimum distance between the rays of cameras observing the same 3D point. Other works incorporate depth into the BA cost function~\cite{melbouci2015bundle,scherer2012using,scherer2013efficient,zhang2021pl}. BA problems are by no means limited with these formulations. Additional constraints might reflect the scene structure for more complex scene models that include semantics, planes, geometric primitives, or objects. For instance, CPA-SLAM~\cite{ma2016cpa} models a scan with a set of planes and penalizes the angle between normals of planes observed from different frames. KDP-SLAM~\cite{hsiao2017keyframe} extracts planes from the fused depth maps, matches these planes iteratively, and penalizes point-to-plane distances for points in the landmark planes. In BAD SLAM, the scan is represented as surfels~--- oriented 3D disks with visual descriptors; similar to CPA-SLAM, the angle between surfel normals is minimized. 

We do not build a special scene representation to enforce geometric consistency in our approach. Instead, we penalize the distance between the matched keypoints backprojected to 3D space using depth maps. Such point-to-point error calculated in 3D space increases the robustness of BA and allows to handle difficult configurations without incurring the risks posed by a reprojection-based cost function. 

\subsection{Semantic-based Pose Refinement}

SLAM methods that estimate and refine camera poses might leverage semantic information in various ways: from ignoring matched keypoints with different semantic labels~\cite{chen2018large} to more inventive object-based approaches. For instance, Frost et al.~\cite{frost2016object} adds a BA term based on the size of detected objects and proves it to prevent scale drift over a long trajectory. Other SLAM methods~\cite{bescos2018dynaslam, brasch2018semantic, li2021dp,xiao2019dynamic} exploit semantic segmentation to remove or detect potential moving objects. In our camera refinement approach, we are interested in detecting structural elements rather than objects. Specifically, we need the semantic labels to create floor and walls point clouds used in refinement.

\subsection{Floorplan-Aware 3D Reconstruction}

The floorplan can facilitate 3D reconstruction in various applications. Howard et al.~\cite{howard2021lalaloc} uses a floorplan-based 3D model for indoor localization and estimates camera pose by comparing image features and layout features calculated on a grid. Wijmans et al.~\cite{wijmans2017exploiting} aligns RGB-D panoramas of large indoor scenes with a floorplan. Goran et al.~\cite{goran2018RBPF} utilizes a floorplan in the grid-based Rao-Blackwellized particle filter and shows that initializing the internal grid with the floorplan information allows obtaining a more precise 2D map of an environment. Contrary to other works, Mielle et al.~\cite{mielle2019autocomplete} does not bind the SLAM map with the floorplan but matches the floorplan onto the SLAM map to complete missing information and unexplored areas.

Rent3D~\cite{liu2015rent3d} takes a floorplan and a set of RGB images as inputs, estimates camera poses, and backprojects pixels onto the generated coarse mesh. This approach provides a non-realistic 3D model with objects projected onto surfaces; moreover, it is limited to one-room scenes. Plan2Scene~\cite{Vidanapathirana2021Plan2Scene} also constructs a 3D model, yet expands to multiple rooms and generates more realistic surfaces via texture synthesis. Either way, Rent3D scans lack furniture, and Plan2Scene replaces scene objects with CAD models. Differently, we use floorplan not to build a 3D model resembling the original scene but to reconstruct an actual scene.

Overall, none of the existing methods address the problem in the same formulation. Since we cannot compare with competing approaches, we analyze each component of our method: we expound the motivation, propose several design choices for this component, and compare these choices quantitatively and qualitatively in ablation studies.

\section{METHOD}

The pipeline of the proposed method is shown in Fig.~\ref{fig:scheme}. Calculating our floorplan-aware BA cost function requires additional steps: converting a floorplan image into a 3D floorplan model, estimating the transformation between the scan and the 3D floorplan, applying semantic segmentation, and constructing floor and walls point clouds. Below, we describe these steps in detail.

\subsection{Scan-to-floorplan Alignment}

\textbf{Finding gravity direction.} Processing a given scan, we first assure that the $y$-axis is pointing upside and the floor surface is horizontal. The gravity direction might be pre-defined, obtained from IMU measurements, or estimated (details can be found in Subsec.~\ref{subsec:implementation_details}).

\textbf{Construction of a boundary scan.} Aligning two non-identical point clouds is the most simple when they are alike. Since the floorplan contains only walls, we remove the floor and furniture from the scan (as described in Subsec.~\ref{subsec:implementation_details}). We refer to the filtered scan as to the \emph{boundary scan} since it represents boundaries between rooms.

\begin{figure}[h!]
    \centering
    \setlength{\tabcolsep}{1pt}
    \begin{tabular}{cc}
    \includegraphics[width=0.49\linewidth]{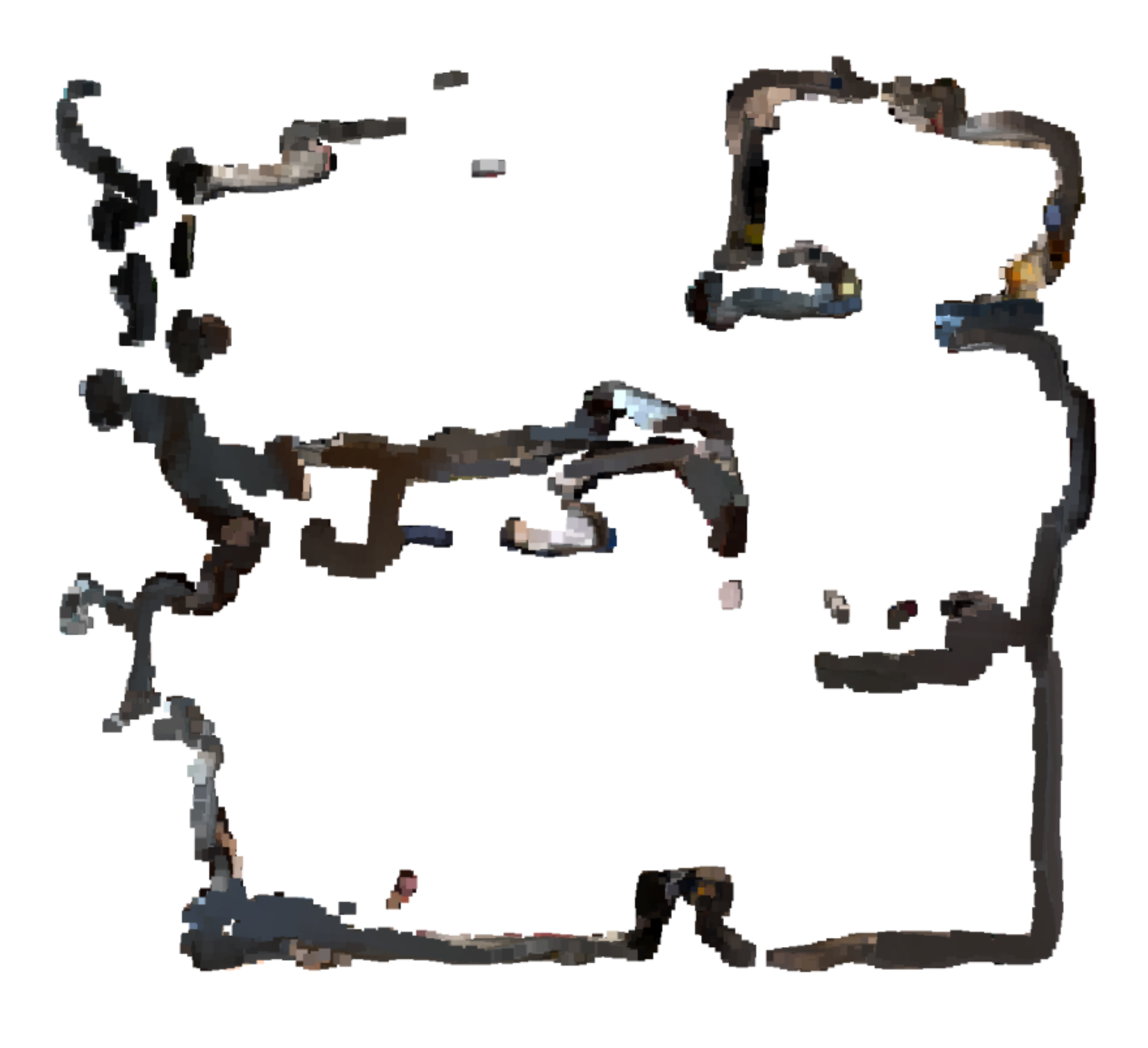} & 
    \includegraphics[width=0.49\linewidth]{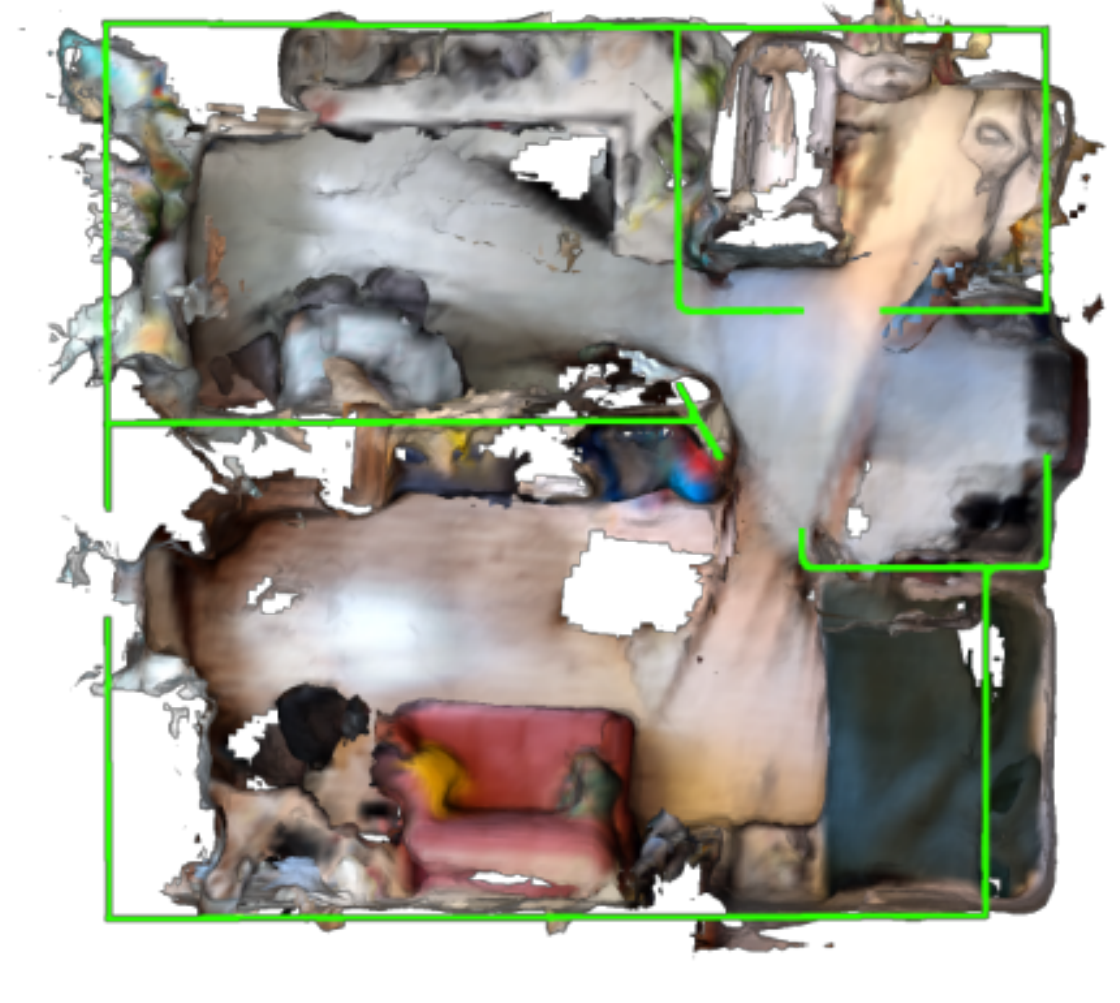} \\
    \end{tabular}
    \caption{The boundary scan (left) and the original scan aligned with a floorplan (right).}
    \label{fig:alignment}
\end{figure}

\begin{figure}[h!]
    \centering
    \includegraphics[width=0.4\linewidth]{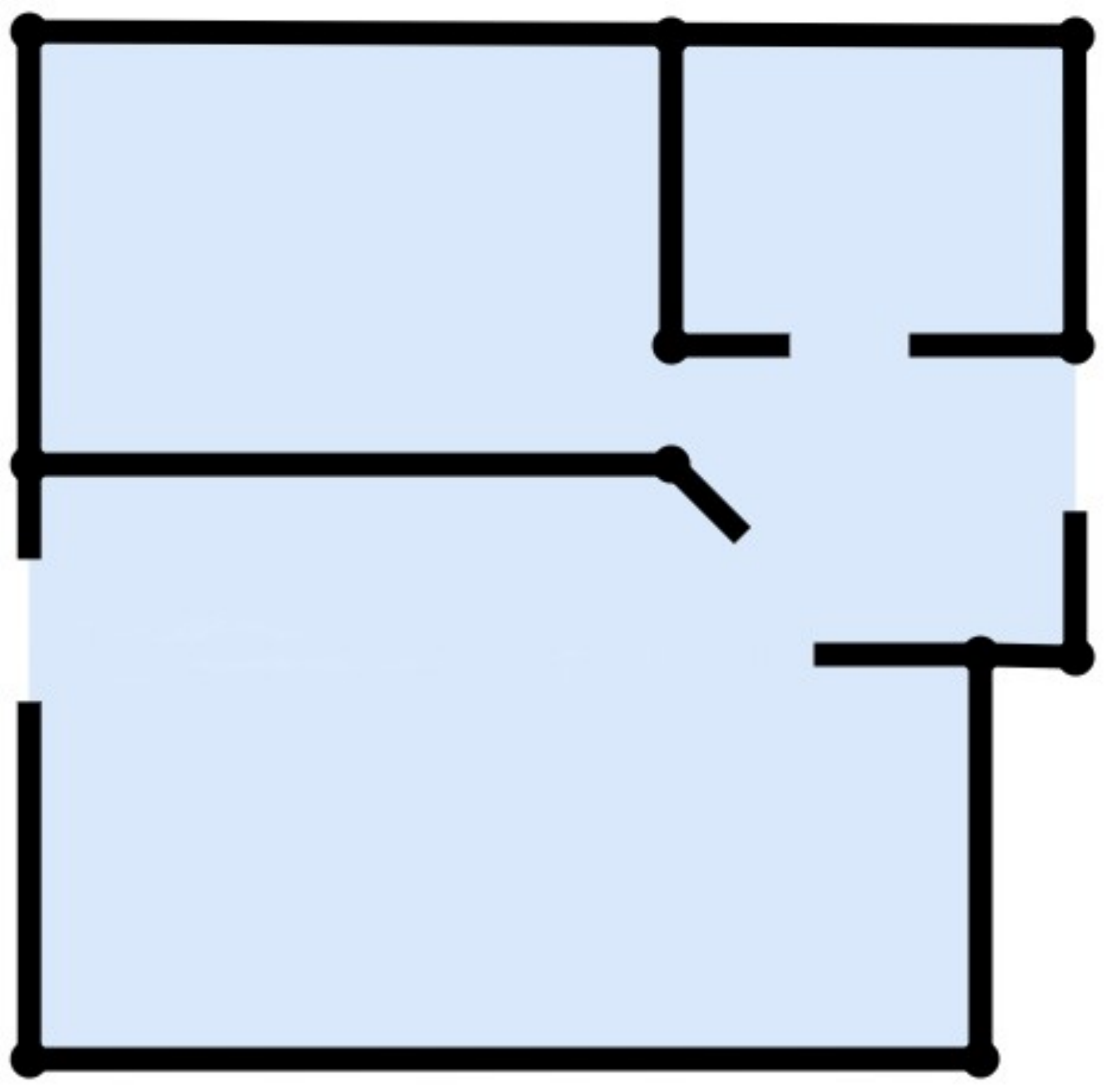}
    \caption{The image of a technical floorplan depicts the general structure of our self-captured environment.}
    \label{fig:floorplan_image}
\end{figure}

\textbf{Creating a 3D floorplan.} We assume that a floorplan is a vector image with walls depicted as segments (Fig.~\ref{fig:floorplan_image}). 

\begin{figure}[ht!]
    \centering
    \setlength{\tabcolsep}{1pt}
    \begin{tabular}{cc}
    \includegraphics[width=0.49\linewidth]{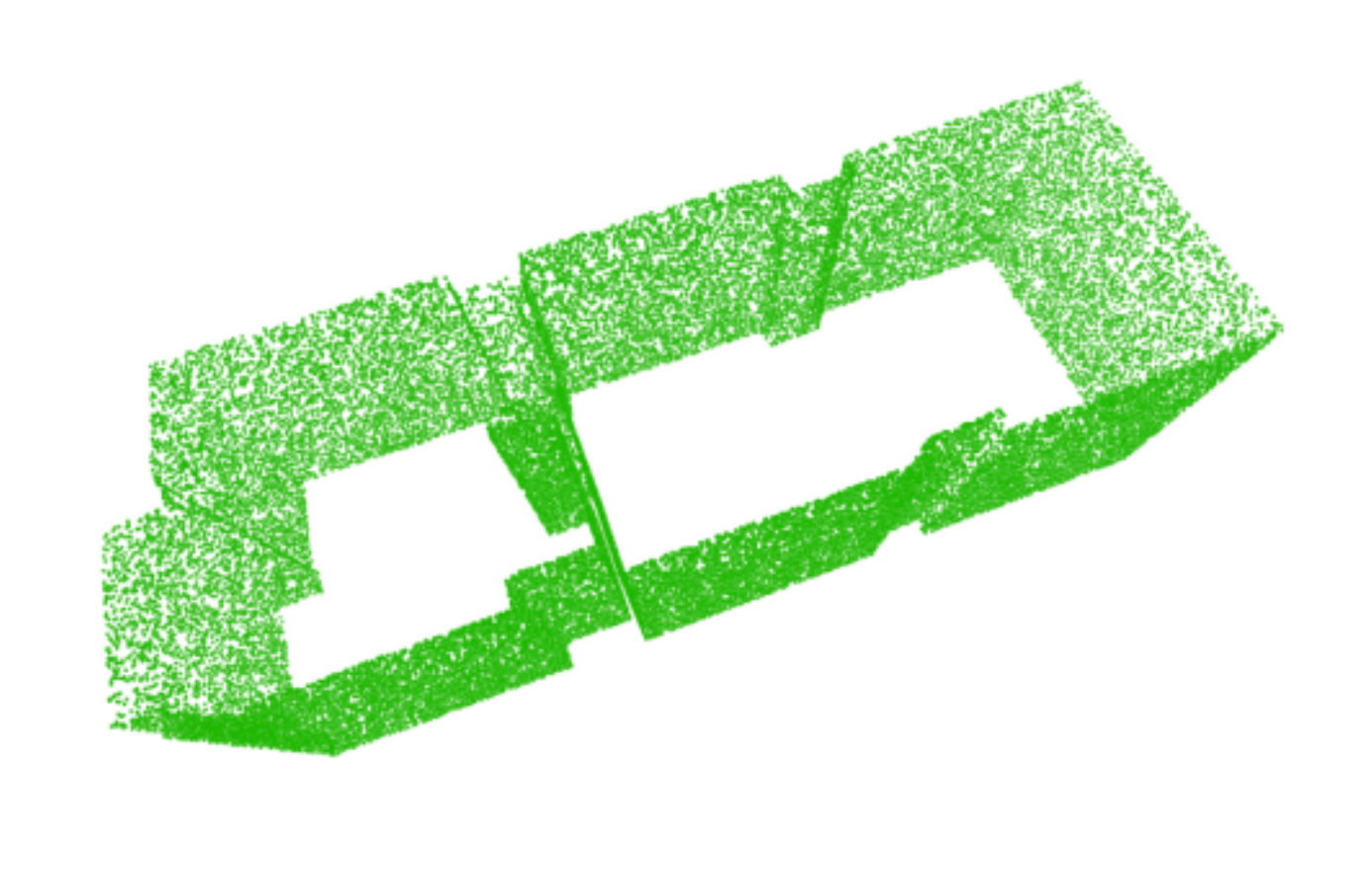} & 
    \includegraphics[width=0.49\linewidth]{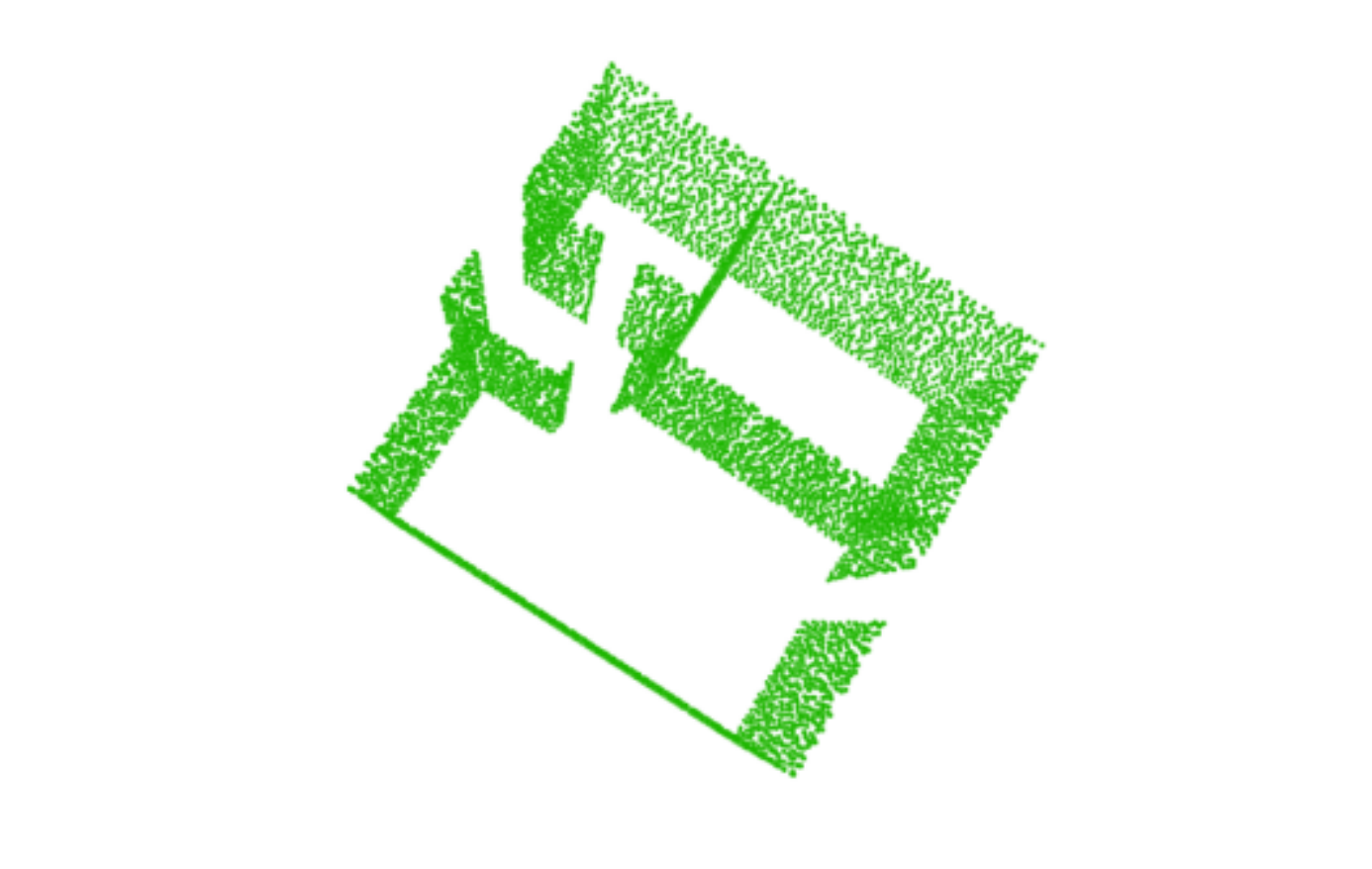}\\
    \end{tabular}
    \caption{The 3D floorplan of a Redwood scan (left) and our self-captured scan. For Redwood, we generate a floorplan by projecting ground truth scan onto the horizontal plane. Our self-captured scans of apartments and office environments are supplied with PNG images of technical floorplans, which we convert into the vector images.}
    \label{fig:3d_floorplan}
\end{figure}

For a 2D floorplan segment with endpoints $(u_1, v_1)$ and $(u_2, v_2)$, we create a 3D rectangle with vertices $(u_1, y_{min}, v_1), \ (u_1, y_{max}, v_1), \ (u_2, y_{min}, v_2)$, and $(u_2, y_{max}, v_2)$, where $y_{min}$ and $y_{max}$ are the minimal and maximal $y$-values of the gravity-aligned scan, respectively. This rectangle approximates (up to shift and scale) the 3D position of the wall depicted as the floorplan segment. To create a 3D wall surface corresponding to this segment, we randomly and uniformly sample points inside this rectangle. By applying the described procedure to each floorplan segment, we obtain a 3D floorplan model (or 3D floorplan for brevity) of the same height as the scan. The examples of such 3D floorplans are visualized in Fig.~\ref{fig:3d_floorplan}.

\textbf{Aligning the boundary scan with the 3D floorplan.} The transformation might be either pre-defined or estimated (the procedure is explicated in Subsec.~\ref{subsec:implementation_details}). The result of the scan-to-floorplan alignment is shown in Fig.~\ref{fig:alignment}. As one might see, the scan is corrupted and requires additional correction.

\subsection{Floor and Walls Extraction}

\begin{figure}[!h]
    \centering
    \begin{subfigure}[t]{0.49\linewidth}
        \centering
        {\setlength{\fboxsep}{0pt}
        \fbox{\includegraphics[width=1.\linewidth]{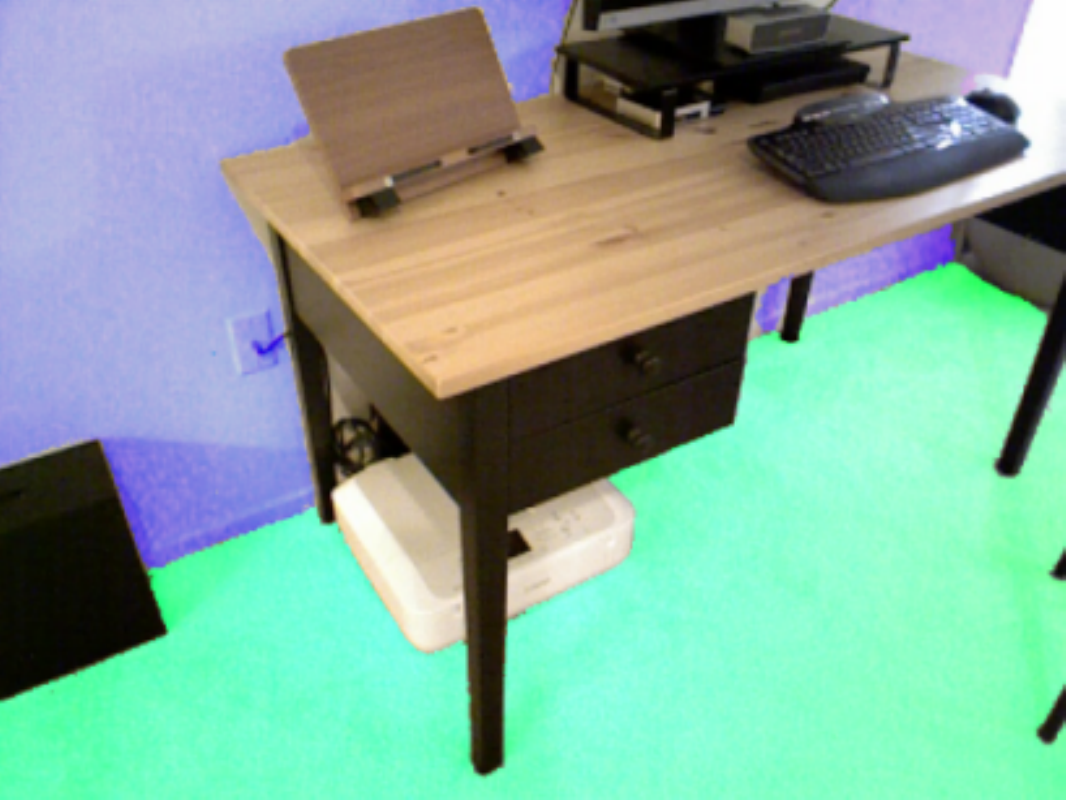}}}
        \caption{ Frame with segmented floor and wall \label{fig:sequence_frame}}
    \end{subfigure}
    \hfill
    \begin{subfigure}[t]{0.49\linewidth}
        \centering
        \includegraphics[width=1.\linewidth]{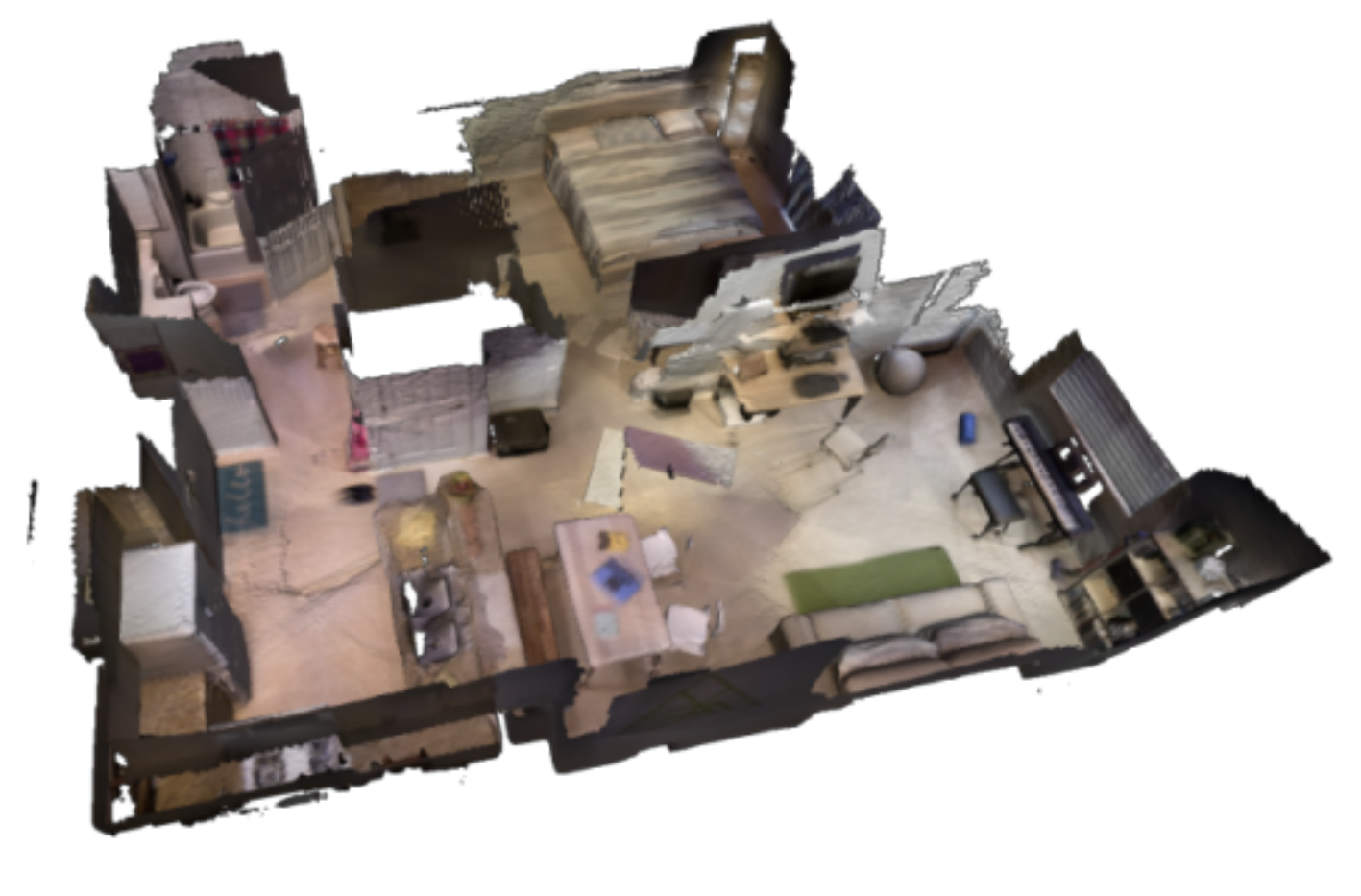}
        \caption{ Point cloud $P$ \label{fig:full_pcd}}
    \end{subfigure}
    \hfill
    \begin{subfigure}[t]{0.49\linewidth}
        \centering
        \includegraphics[width=1.\linewidth]{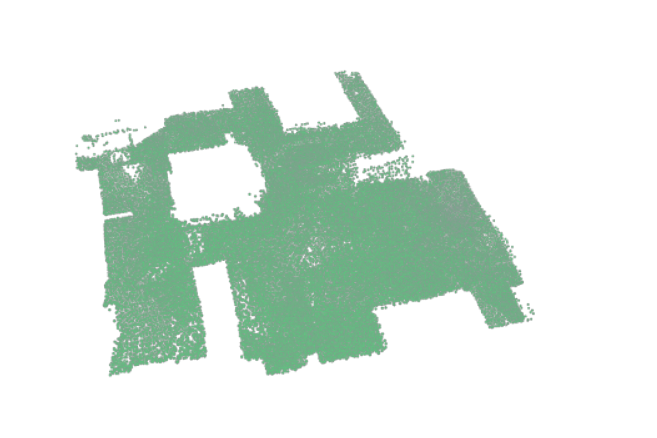}
        \caption{ Floor point cloud $P_F$ \label{fig:floor_pcd}}
    \end{subfigure}
    \hfill
    \begin{subfigure}[t]{0.49\linewidth}
        \centering
        \includegraphics[width=1.\linewidth]{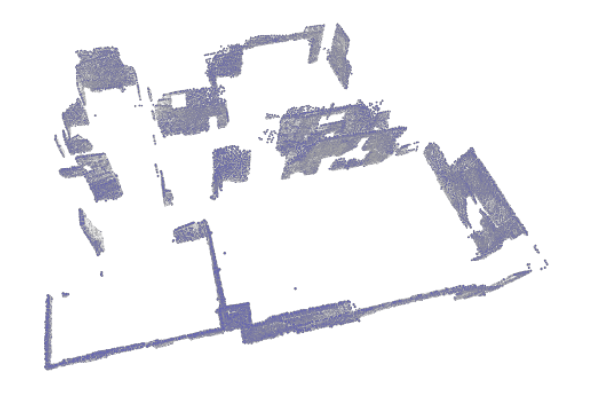}
        \caption{  Walls point cloud $P_W$ \label{fig:wall_pcd}}
    \end{subfigure}
    \caption{Floor and walls are segmented in each RGB frame~(\ref{fig:sequence_frame}) and backprojected to 3D space, giving floor~(\ref{fig:floor_pcd}) and walls point clouds~(\ref{fig:wall_pcd}), respectively.}
    \label{fig:floor_wall_PCD}
\end{figure}

To find floor and walls in the scan $P$, we apply 2D semantic segmentation~\cite{MSeg_2020_CVPR} for each RGB image. The points labeled as \textit{floor}, \textit{ground}, or \textit{carpet} are backprojected to 3D space using depth maps; altogether they comprise a floor point cloud $P_F$. Similarly, walls point cloud $P_W$ is constructed from the points classified as \textit{wall}. The original scan $P$, the floor point cloud $P_F$ and the walls point cloud $P_W$ are depicted in Fig.~\ref{fig:floor_wall_PCD}.

\subsection{Our BA Cost Function}

\begin{figure*}[t]
\centering
\setlength{\tabcolsep}{10pt}
\begin{tabular}{cc}
    \includegraphics[width=0.35\linewidth]{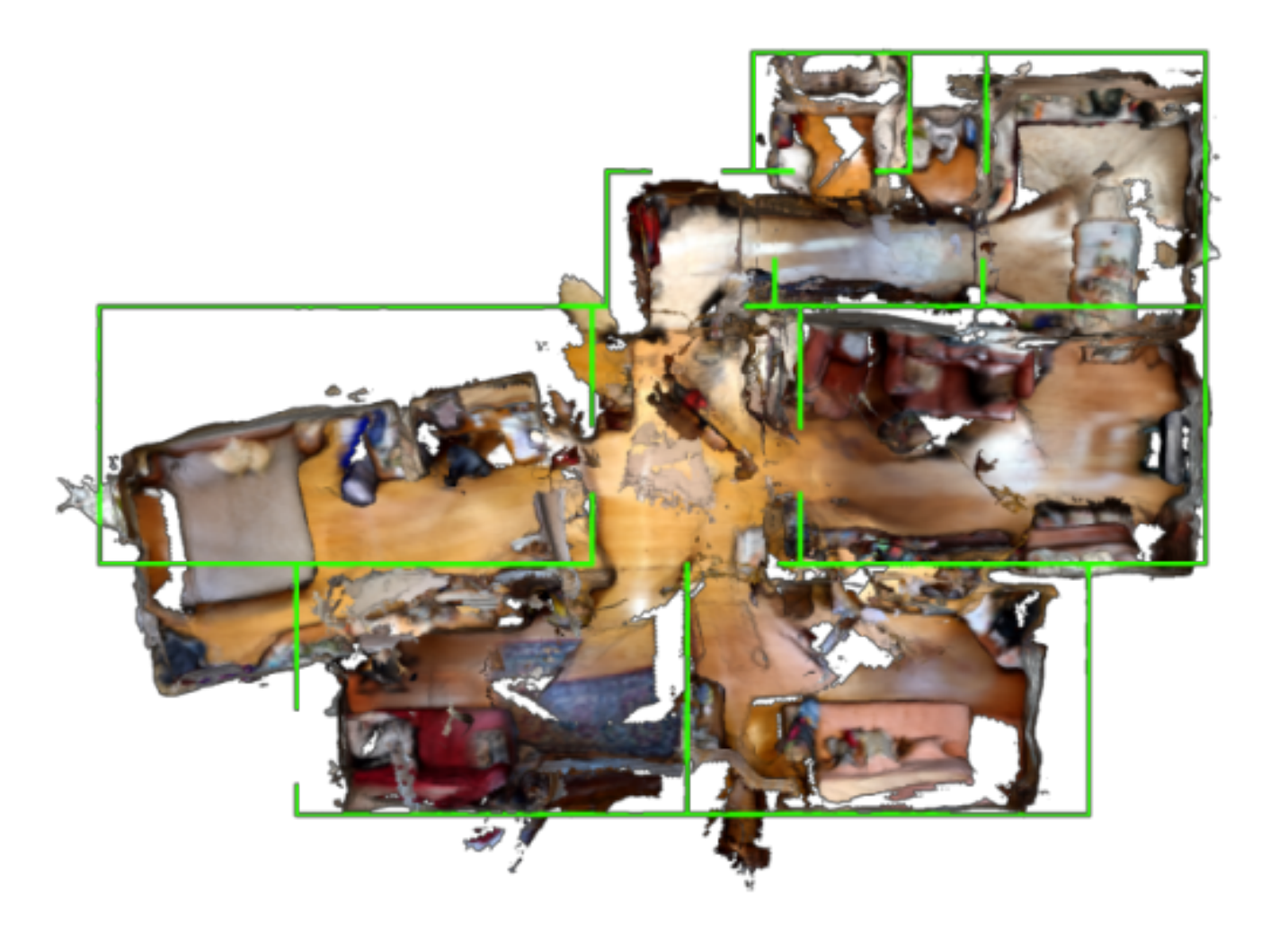} &
    \includegraphics[width=0.35\linewidth]{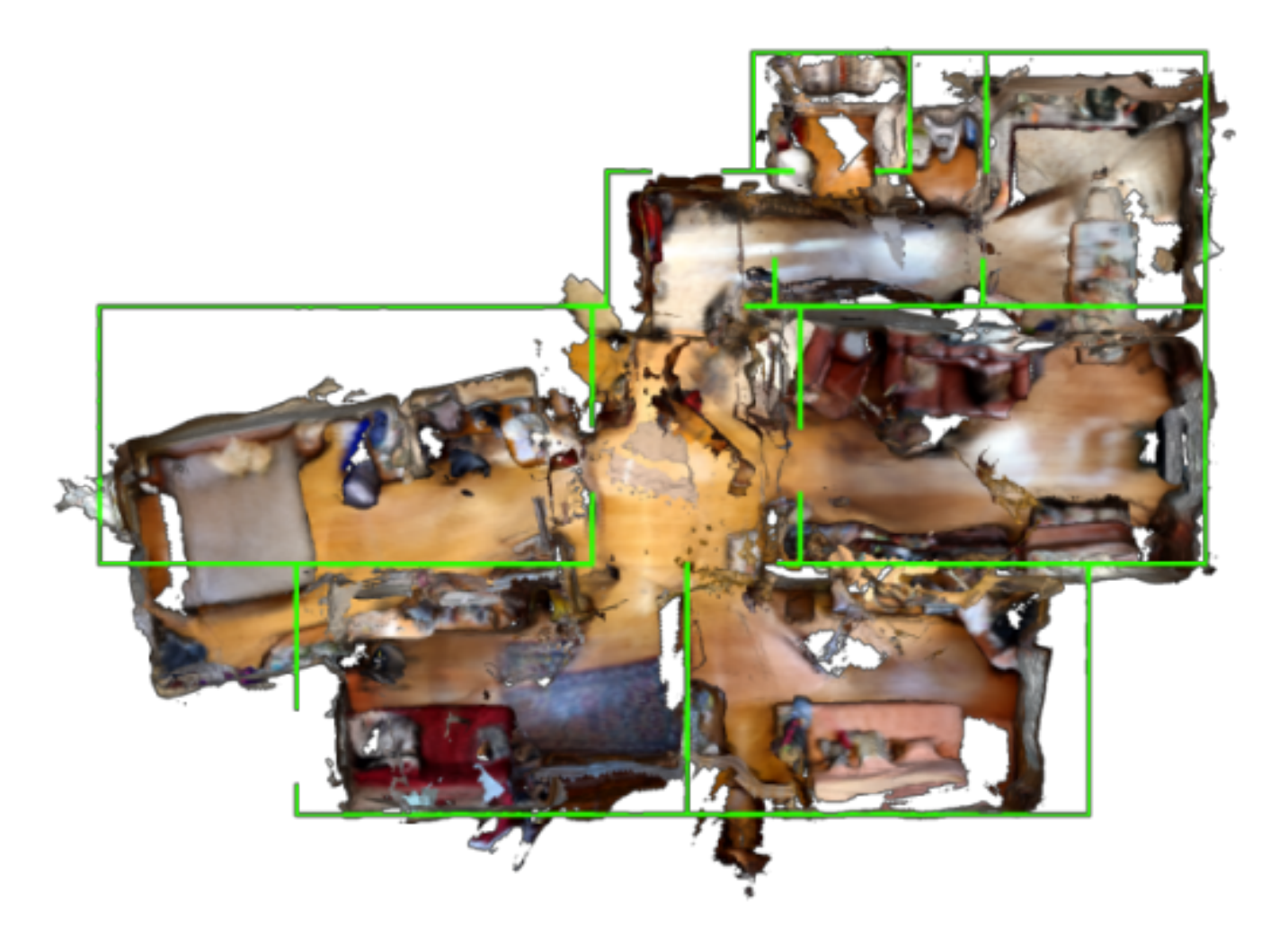} \\
    (a) Raw scan & (b) Floor-to-plane pulling \\
    \includegraphics[width=0.35\linewidth]{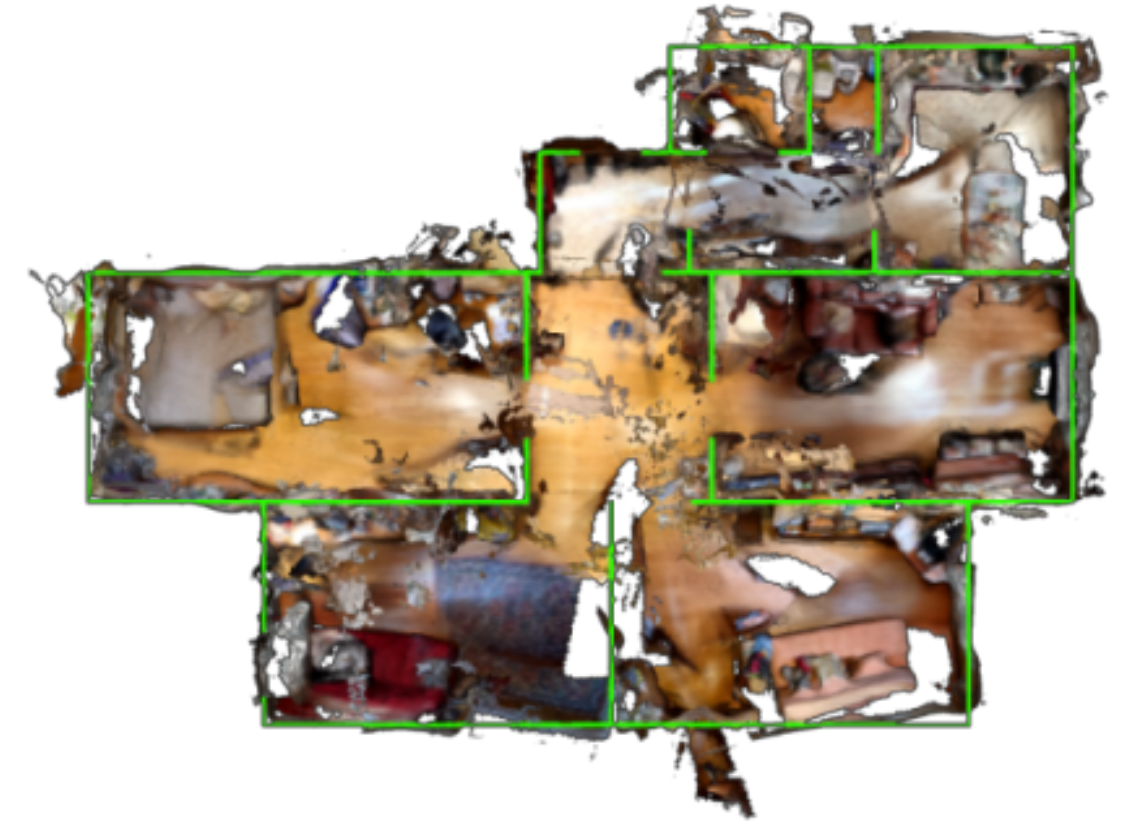} &
    \includegraphics[width=0.35\linewidth]{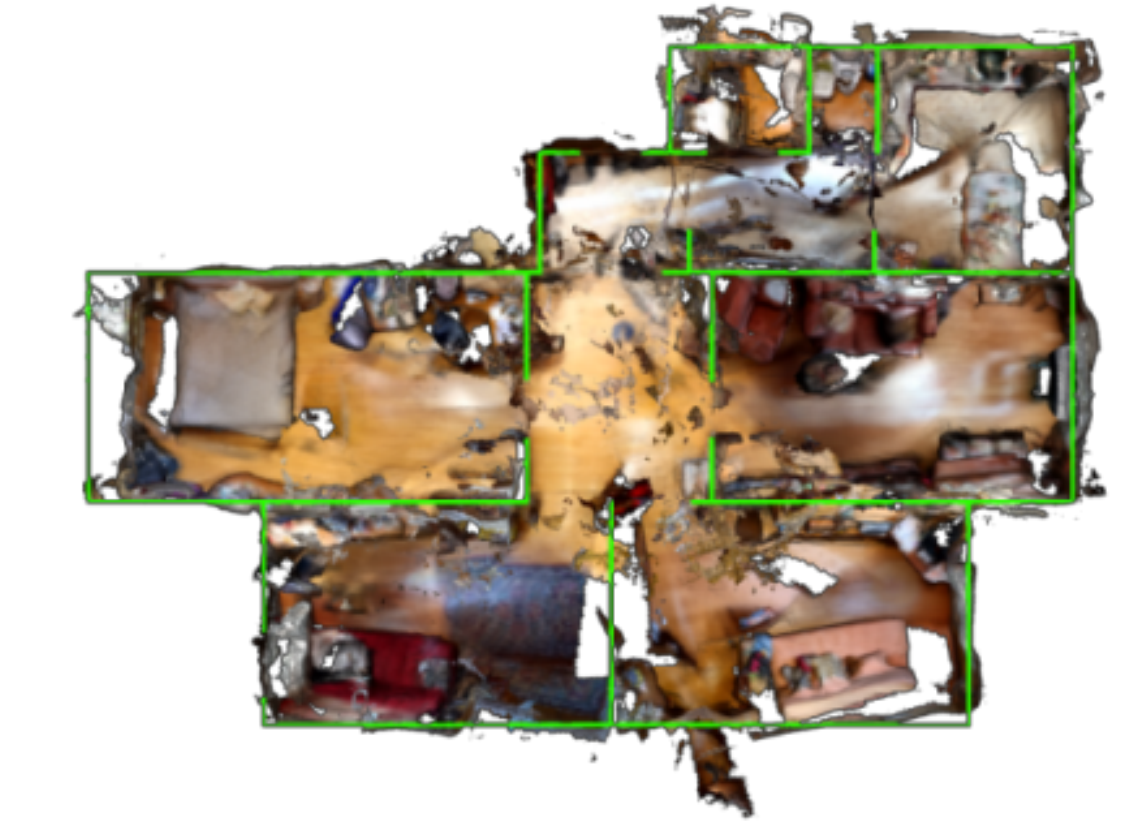} \\
    (c) Walls-to-floorplan pulling & (d) Floor-to-plane + walls-to-floorplan pulling \\
\end{tabular}
\caption{Due to an accumulated trajectory drift in our self-captured scan, one of the rooms is misplaced (a). Floor-to-plane pulling (b) reduces the discontinuities. After the walls-to-floorplan pulling (c), the room is placed correctly but the furniture is slightly corrupted. A combination of floor-to-plane and walls-to-floorplan pulling (d) improves the interior reconstruction while allowing to restore the general scene structure.}
\label{fig:smapper_rechnoy}
\end{figure*}

We refine camera poses by minimizing the three-term cost function: the \emph{geometric term} focuses on the reconstruction consistency, the \emph{floor term} aims at making the floor surface flat, and the \emph{walls term} penalizes the discrepancy between the reconstructed walls and the floorplan:
$$
L = L_{geom} + \lambda_{floor}L_{floor} + \lambda_{walls}L_{walls}.
$$

\textbf{Geometric term.} To enforce structural consistency in 3D space, we employ a geometric term representing the 3D discrepancy of 3D point estimates from two views. Hereinafter, we refer to this term as to 3D point error. 

To obtain 3D points, we extract 2D keypoints from all given frames with SuperPoint~\cite{superpoint} and match these keypoints across the frames. We backproject the keypoints to 3D space using depth maps and calculate the distances between 3D points backprojected from different frames.

For a $i$-th frame $I_i$, let $K_i$ be the set of keypoints detected in this frame. For a keypoint $k \in K_i$, let $p(k)$ denote it backprojection to 3D space with respect to the $i$-th camera. 
For each pair of matched keypoints $\left(k, k' \right)$, we backproject these keypoints, which gives a pair of 3D points $\left( p, p' \right)$. Denoting all such pairs as $M$, we formulate our geometric term as:
$$
L_{geom} = \sum_{(p, p') \in M} \lVert p - p'\rVert_2.
$$

\subsection{Floor-to-Plane Pulling}

The floor term prevents the floor from being uneven and layered by pulling its points to the floor plane $\pi_F$. $\pi_F$ is fitted to the $P_F$ points once and remains fixed during the optimization.
The floor term is calculated as the mean distance between $P_F$ and $\pi_F$:
$$
L_{floor} = \sum_{p\in P_F} dist(p, \pi_F).
$$

\subsection{Walls-to-Floorplan Pulling}

In a reconstructed scan, walls typically consist of numerous plane segments. Ideally, these segments are flat and aligned with the floorplan walls. Below, we describe three walls-to-floorplan pulling strategies that target these requirements: the nearest point, the iterative nearest wall, and the fixed nearest wall strategy. 

\textbf{Nearest point.}
For each point $p \in P_W$, we find the nearest 3D floorplan point $q(p)$ and calculate the distance between these two points. By averaging such distances, we obtain the following walls term: 
$$
L_{walls}^{NP} = \sum_{p\in P_W} \lVert p - q(p) \rVert_2.
$$
The point-to-point pulling brings the wall points closer to the floorplan; however, it does not make the walls any flatter.

\textbf{Iterative nearest wall.} 
Alternatively, the $P_W$ points can be pulled not to the nearest 3D floorplan points but to the planes. In this approach, the 3D floorplan is represented as a set of planes $\left\{\pi^f \right\}$ approximating the walls. For each point $p \in P_W$, we find its nearest 3D floorplan point $q(p)$ and estimate the distance to the corresponding wall plane $\pi^f(q(p))$. We average these distances to obtain the walls term:
$$
L_{walls}^{INW} = \sum_{p\in P_W}dist\left(p, \pi^f(q(p))\right).
$$

While being more stable than the nearest point strategy, the iterative nearest wall strategy still has serious drawbacks. First, finding the nearest points on each step makes this strategy very slow. Second, two points in the $P_W$ cloud belonging to the same wall may be pulled to different 3D floorplan planes, especially in the wall intersection areas.

\textbf{Fixed nearest wall.} 
To speed up optimization, we establish the correspondence between $P_W$ points and the 3D floorplan wall planes and keep it unchanged during optimization. 

Specifically, we cluster $P_W$ points according to their normals and fit vertical wall planes. Then, we match wall planes with the 3D floorplan wall planes, assuming that the wall plane $\pi$ and the 3D floorplan wall plane $\pi^f$ correspond to each other if they are mutually nearest and parallel. We match each point $p \in P_W$ belonging to the wall plane $\pi$ with the 3D floorplan wall plane $\pi^f$ corresponding to $\pi$. Hence, we optimize over a set $\Pi$ of ($p$, $\pi$), where $\pi$ is estimated once using the initial location of $p$ and fixed during optimization. We calculate a point-plane distance for each pair in $\Pi$ and summarize such distances to obtain the walls term: 
$$
L_{walls}^{FNW} = \sum_{(p, \pi) \in \Pi}dist\left(p, \pi\right).
$$
The established correspondences depend on the mutual assignment of points in $P_W$ and floorplan wall planes, preventing two close-by points from being assigned to different planes and then pulled in different directions. In addition, when the correspondences are fixed, only the point-plane distances are estimated at each optimization step, making this strategy much faster than the other two.

Since the matches between the walls and the floorplan walls may change during optimization, we alternate the loss minimization and alignment steps.

\section{EXPERIMENTS}

\subsection{Datasets}

Given the novelty of the floorplan-aware camera pose optimization problem, it is necessary to formulate an original evaluation protocol including data acquisition, data pre-processing, and metrics calculation. Our approach can be benchmarked on the posed RGB-D data containing floorplans. Accordingly, most RGB-D SLAM datasets are directly inapplicable due to the absence of ground truth floorplans.

\textbf{Redwood.}
To the best of our knowledge, Redwood~\cite{redwood} is the only dataset suitable for our experiments. Redwood contains 5 posed RGB-D sequences with 25k frames on the average recorded via Asus Xtion Live camera, and provides dense ground truth scans acquired with FARO Focus 3D X330 HDR scanner. Accordingly, we can extract a floorplan given ground truth scans; such a floorplan would not depend on the RGB-D measurements and tracking results. Besides, Redwood scenes have multiple rooms and many partitions. Handling such data is challenging for SLAM methods, which opens up possibilities for further improvements.

The original Redwood camera poses are obtained via a tracking algorithm. Tracking methods are developing rapidly, so we re-estimate camera poses with a recent DROID-SLAM~\cite{teed2021droid} that achieves state-of-the-art accuracy while experiencing fewer failures than competing approaches. 

\begin{table*}[h!]
\caption{Results obtained with different combinations of geometric, floor, and walls BA terms.}
\label{tab:quantitative_comparison}
\begin{center}
\begin{tabular}{|l|c|c|c|c|c|c|c|c|c|}
    \hline
    Dataset & Geometric term & Floor term & Walls term & MME\textdownarrow & MPV\textdownarrow & MOM\textdownarrow & NND\textdownarrow & NSD\textdownarrow\\
    \hline
    \multirow{5}{*}{Redwood} & - & - & - & -3.81 & 18.68 & 33.07 & 9.43 & 8.37\\
    & + & - & - &  -3.84 & 17.98 & 32.63 & 9.09 & 8.49\\
    & + & + & - & -3.99 & 14.72 & 29.23 & 8.48 & 8.97 \\
    & + & - & + & -3.94 & 15.52 & 29.61 & 8.62 & 6.53\\
    & + & + & + & \textbf{-4.02} & \textbf{14.46} & \textbf{28.94} & \textbf{8.38} & \textbf{6.31}\\
    \hline
    \multirow{5}{*}{Self-captured data} & - & - & - & -3.53 & 30.01 & 81.55 & N/A & 28.23\\
    & + & - & - & -3.61 & 25.51 & 65.19 & N/A & 28.17 \\
    & + & + & - & -3.62 & 25.01 & 61.62 & N/A & 28.49 \\
    & + & - & + & -3.61 & 25.13 & 61.84 & N/A & 25.14\\
    & + & + & + & \textbf{-3.63} & \textbf{24.86} & \textbf{59.08} & N/A & \textbf{25.10}\\
    \hline
\end{tabular}
\end{center}
\end{table*}

\textbf{Self-captured data.} We also test our approach using our own data: 5 scans of multi-room apartments and office environments supplied with a technical floorplan. These scans are captured with a Samsung Galaxy S20+ smartphone equipped with a ToF depth sensor and Google ARCore tracking system. Google ARCore is widely used in mobile AR applications to facilitate real-time inference. However, we observe that it accumulates errors over time when used for recording long trajectories.

\subsection{Metrics}

We validate our approach using metrics of two different types. The reconstruction metrics (MME, MPV, MOM, NND) assess the quality of the point cloud obtained via backprojection. The NSD metric measures the difference between scan and floorplan walls.

\textbf{MME~\cite{MME_MPV}.} For each point in a reconstructed scan, we calculate the entropy of points within a given radius. Averaging these values gives the Mean Map Entropy, or MME.

\textbf{MPV~\cite{MME_MPV}.} For the Mean Plane Variance (MPV), we assume that most scan surfaces are planar. For each point, we select points within a given radius, approximate a plane from these points, and find the distance of every point to this plane. MPV is an average variance of these distances.

\textbf{MOM~\cite{kornilova2021benchmark}.} Mutually Orthogonal Metric, or MOM, is also based on the point-to-plane distance variance. Unlike MPV, it is estimated only for the points located on three orthogonal planes that are established for each depth map independently. 

\textbf{NND.} As Redwood contains ground truth scans, we estimate Nearest Neighbor Distance (NND) in addition to the no-reference metrics (MME, MPV, MOM). We align reconstructed scans with ground truth ones via Deep Global Registration~\cite{choy2020deep}, then calculate NND as an average distance from the points of a reconstructed scan to their nearest points of a ground truth scan. 

\textbf{NSD.} We also measure the discrepancy between the scan and the floorplan with Nearest Segment Distance (NSD), which is the average distance between $P_W$ points and their nearest floorplan segments.

\subsection{Implementation Details}
\label{subsec:implementation_details}

\textbf{Finding gravity direction.} To estimate the gravity direction, we project the scan normals onto a sphere. We assume that the most common vector defines the gravity direction.

\textbf{Construction of a boundary scan.} We remove the floor and furniture from the scan to make it resemble a floorplan. To remove the floor, we build the histogram of the $y$-coordinates of the scan points. The peak value of the histogram corresponds to the floor plane, so we remove the points located below a given threshold. Filtering out the points whose horizontal projections are statistical outliers allows us to remove the furniture.

\textbf{Aligning the boundary scan with the 3D floorplan.} We formulate the aligning transformation as a combination of the rotation around the $y$-axis, the shift, and the scale.

First, we estimate the rotation around the $y$-axis based on normals of the boundary scan: we project them onto a sphere and select the three most frequent orthogonal directions. One of these directions coincides with the gravity direction; we project the two other vectors onto the horizontal plane. Similarly, we find the horizontal projection of the most common normal in the 3D floorplan. Then, we rotate the 3D floorplan so that this projected normal is aligned with any of two projected boundary scan normals or directed oppositely (totalling 4 rotations). For each rotation, we estimate the average distance between points in the boundary scan and their nearest points in the rotated 3D floorplan and select the rotation with the smallest distance.

The scale is set according to the ratio of $x$-range and $z$-range of the boundary scan and the 3D floorplan. $y$-values are not affected, so the 3D floorplan height remains equal to the scan height. Finally, the shift is chosen so that the boundary scan and the 3D floorplan's geometrical centers coincide.


\textbf{BA.} We implement BA using PyTorch~\cite{pytorch2019paszke}. The loss function is minimized via gradient descent with momentum. The initial learning rate is $10^{-3}$. After $20000$ gradient steps, the learning rate is set to $10^{-4}$, and the process continues until the difference of loss value between iterations becomes less than $10^{-5}$.
We use $\lambda_{floor} = 10, \lambda_{walls} = 0.6$ for Redwood, $\lambda_{floor} = 10, \lambda_{walls} = 0.5$ for our self-captured data. 

\section{RESULTS}

To validate the proposed approach, we try different combinations of the BA terms described above and report quantitative results for Redwood and self-captured data. We also perform ablation studies of geometric and walls terms.

\subsection{Quantitative Results}

According to Tab.~\ref{tab:quantitative_comparison}, BA with only a geometric term improves the reconstruction quality. The accuracy gain is the most significant for the self-captured scans containing inaccurate Google ARCore camera poses. The refined Redwood camera poses also provide better reconstructions than initial estimates obtained with DROID-SLAM. Predictably, walls-to-floorplan pulling significantly improves NSD. The best NSD is achieved when using both pulling terms; however, the effect of the floor term is minor.

\subsection{Qualitative Results}

We refine camera poses using different combinations of the floor-to-plane and walls-to-floorplan pulling and visualize the reconstructed scans. As one might observe, the walls-to-floorplan pulling allows to restore the general scene structure correctly (Fig.~\ref{fig:smapper_rechnoy}). When combining walls-to-floorplan pulling with the floor-to-plane pulling, we obtain the most accurate reconstructions.

\begin{figure}[h!]
\centering
\setlength{\tabcolsep}{1pt}
\begin{tabular}{c}
    \includegraphics[width=0.95\linewidth]{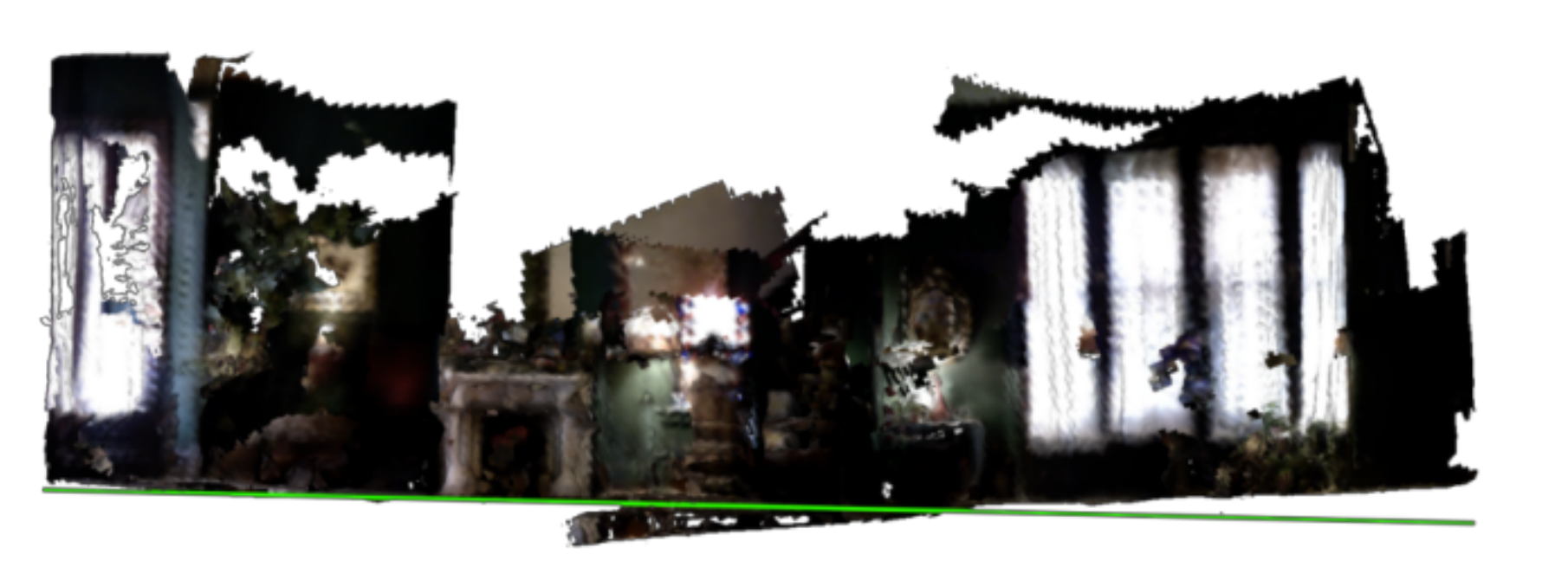} \\
    (a) Before optimization \\
    \includegraphics[width=0.95\linewidth]{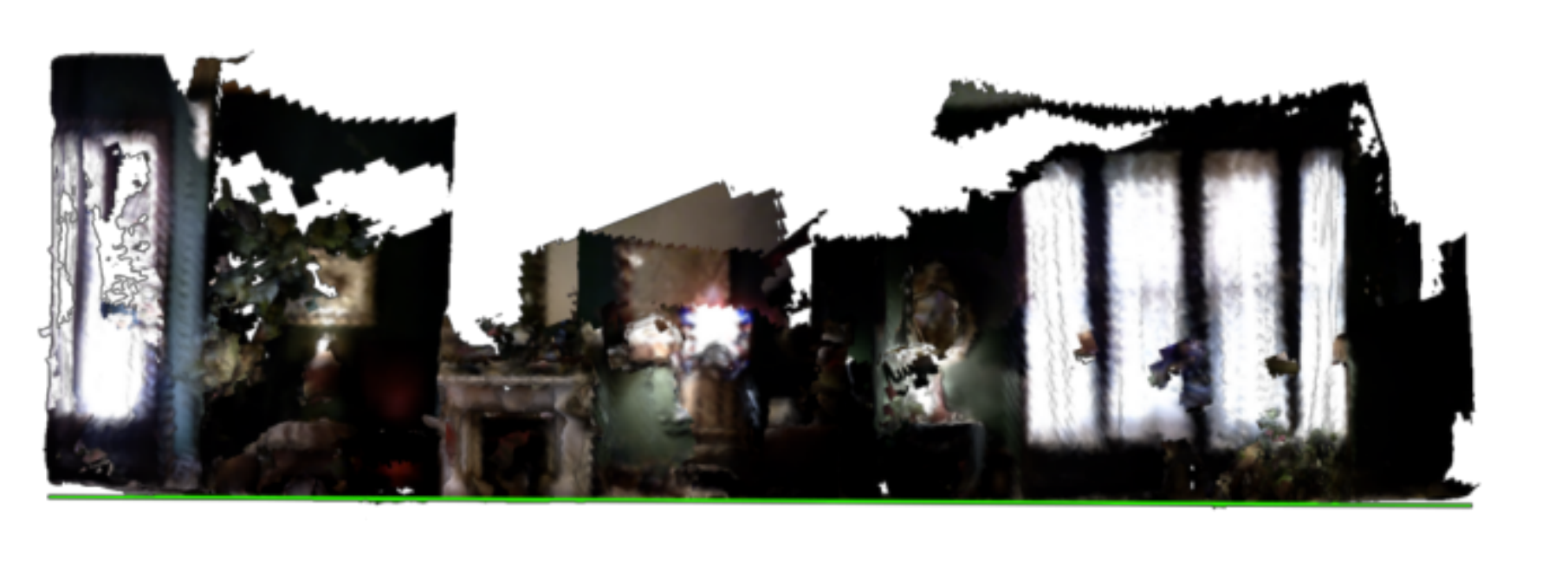} \\
    (b) After optimization \\
\end{tabular}
\caption{Optimization makes the uneven and layered floor of a Redwood scan flat and smooth.}
\label{fig:redwood_lobby}
\end{figure}

\begin{table*}[h!]
\caption{Results of ablation study of different walls-to-floorplan pulling strategies.}
\label{tab:ablation_walls_term}
\begin{center}
\begin{tabular}{|l|l|c|c|c|c|c|c|}
    \hline
    Dataset & Walls-to-floorplan pulling strategy & MME\textdownarrow & MPV\textdownarrow & MOM\textdownarrow & NSD\textdownarrow & Time, ms \\
    \hline
    \multirow{4}{*}{Redwood} & None & -3.81 & 18.68 & 33.07 & 8.37 & -\\
    & Nearest Point & -3.93 & 16.88 & 31.04 & \textbf{4.84} & 66.52 \\
    & Iterative Nearest Plane & \textbf{-3.96} & \textbf{15.39} & 30.91 & 6.31 & 70.47 \\
    & Fixed Nearest Plane & -3.94 & 15.52 & \textbf{29.61} & 6.53 & 6.90\\
    \hline
    \multirow{4}{*}{Self-captured data} & None & -3.53 & 30.01 & 81.55 & 28.23 & - \\
    & Nearest Point & -3.59 & 25.45 & 63.81 & \textbf{16.60} & 66.52 \\
    & Iterative Nearest Plane & -3.60 & 25.50 & 63.88 &  24.18 & 70.47 \\
    & Fixed Nearest Plane & \textbf{-3.61} & \textbf{25.13} & \textbf{61.84} & 25.14 & 6.90\\
    \hline
\end{tabular}
\end{center}
\end{table*}

\begin{table*}[h!]
\caption{Results of ablation study of different geometric terms.}
\label{tab:ablation_geometric_term}
\begin{center}
\begin{tabular}{|l|l|c|c|c|c|}
    \hline
    Dataset & Geometric term & MME\textdownarrow & MPV\textdownarrow & MOM\textdownarrow & NSD\textdownarrow \\
    \hline
    \multirow{6}{*}{Redwood} & None & -3.81 & 18.68 & 33.07 & 8.37 \\
    &  3D point error & \textbf{-4.02} & \textbf{14.46} & \textbf{28.94} & \textbf{6.31}  \\
    & Reprojection error  & -3.91 & 16.94 & 30.78 & 6.96  \\
    & Minimum ray distance & -3.88 & 17.86 & 31.07 &  8.11 \\
    & IBA with 3D point error  & -3.89 & 17.45 & 29.26 &  7.16 \\
    & IBA with reprojection error & -3.88 & 17.73 & 30.55 &   7.45  \\
    \hline
    \multirow{6}{*}{Self-captured data} & None & -3.53 & 30.01 & 81.55 & 28.23\\
    & 3D point error & \textbf{-3.63} & \textbf{24.86} & \textbf{59.08} & \textbf{25.10} \\
    & Reprojection error  & -3.56 & 27.21 & 64.79 & 25.99  \\
    & Minimum ray distance & -3.54 & 29.64 & 78.78 & 28.22  \\
    & IBA with 3D point error  & -3.62 & 25.39 & 63.36 &  25.92 \\
    & IBA with reprojection error  & -3.58 & 27.12 & 64.35 & 26.40  \\
    \hline
\end{tabular}
\end{center}
\end{table*}

\begin{figure}[h!]
\centering
\setlength{\tabcolsep}{1pt}
\begin{tabular}{c}
    \includegraphics[width=0.70\linewidth]{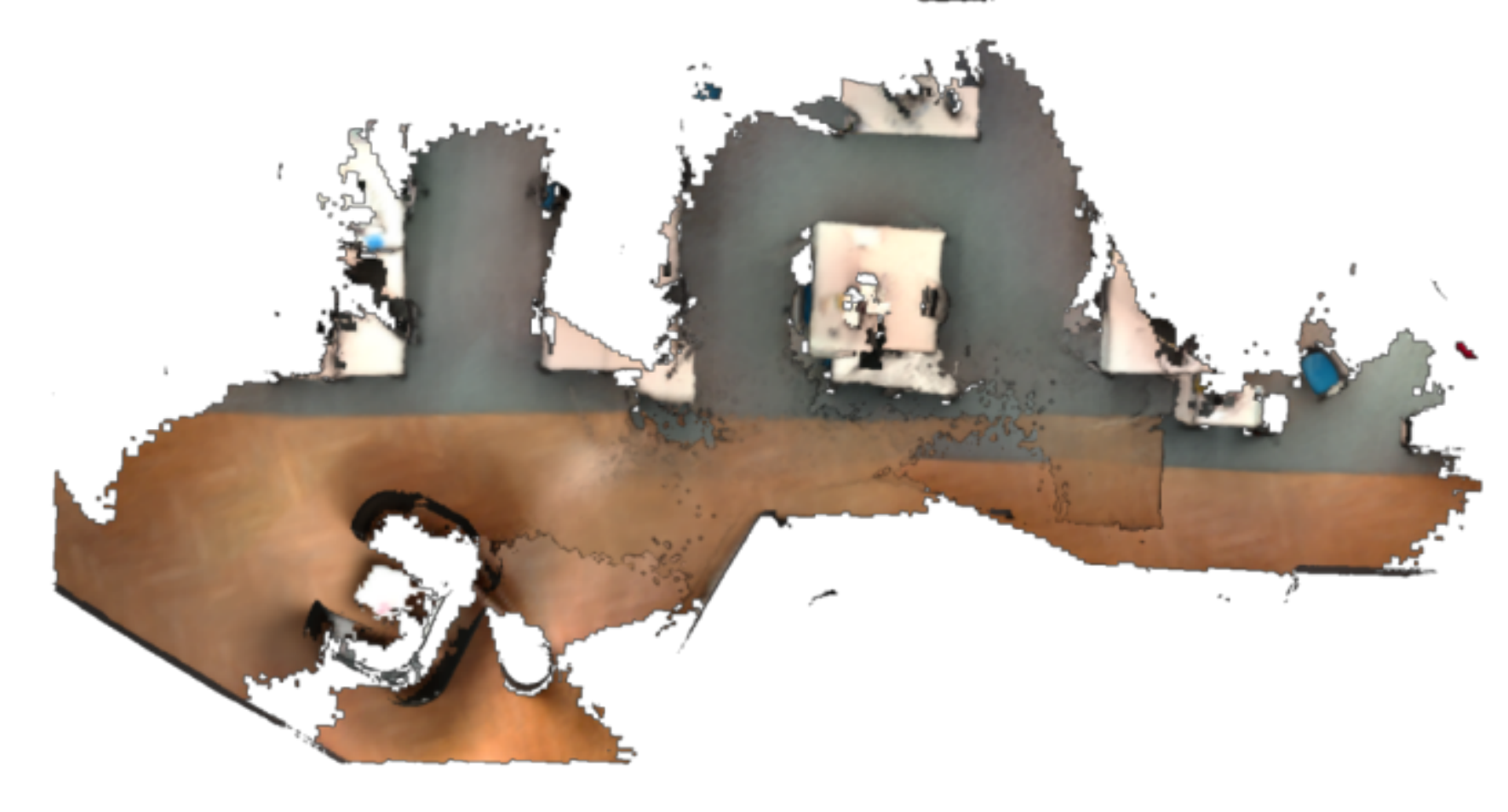} \\
    (a) Before optimization \\
    \includegraphics[width=0.70\linewidth]{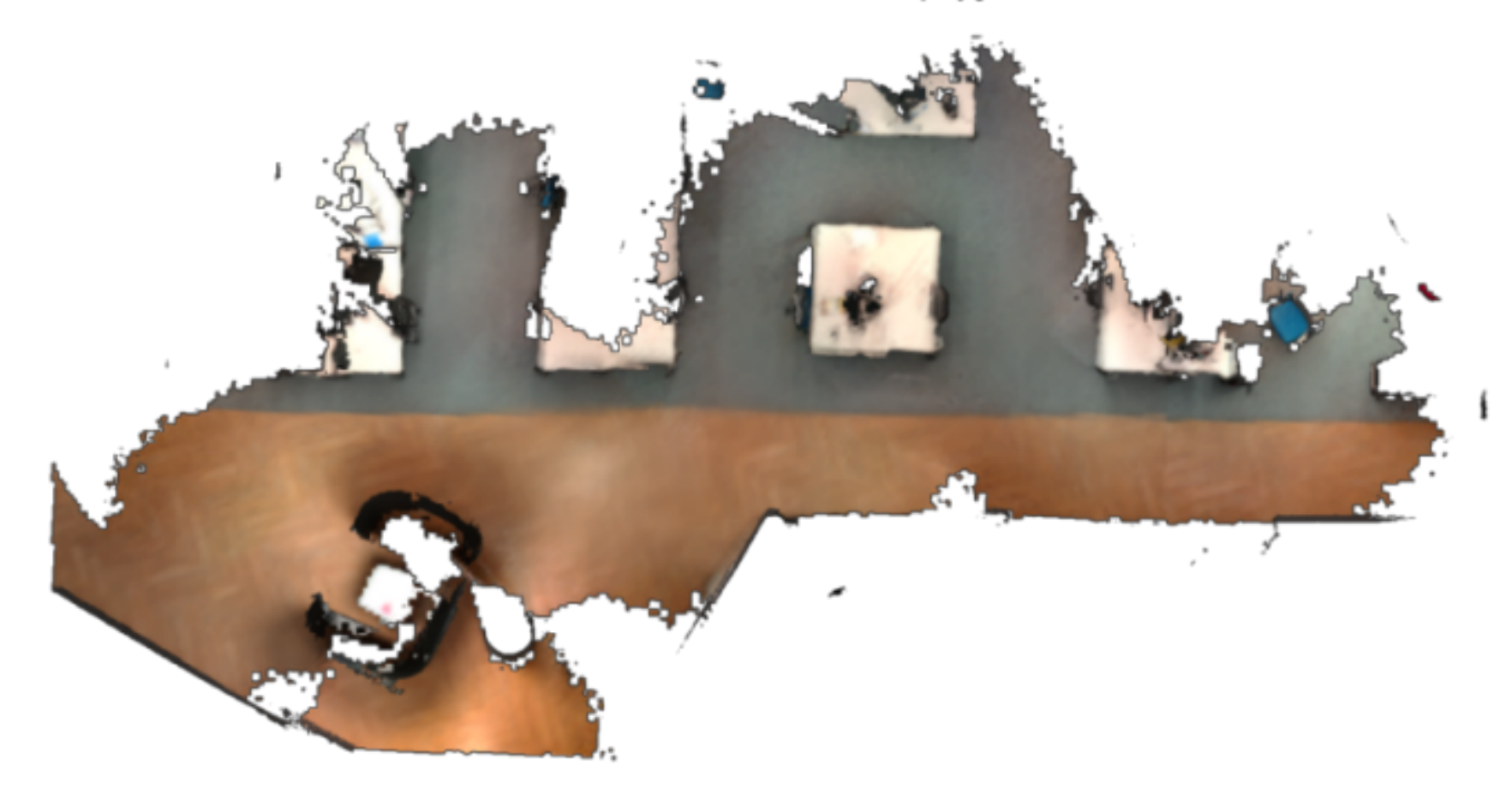} \\
    (b) After optimization \\
\end{tabular}
\caption{After optimization, the floor and furniture in a self-captured scan look less corrupted.}
\label{fig:smapper_office}
\end{figure}

Fig.~\ref{fig:redwood_lobby} demonstrates that floor-to-plane pulling flattens the floor, which is reflected in smaller MME and MPV values. At the same time, walls-to-floorplan pulling allows obtaining a more accurate reconstruction (Fig.~\ref{fig:smapper_office}). Using both pulling terms jointly might impose the risk of the floor and walls being pulled in different directions causing the scan corruption. However, our empirical study disproves this: on the contrary, combining floor and walls terms provides the best results, so we assume these terms are complementary.

\subsection{Ablation studies}

\textbf{Geometric term.} We investigate several formulations of geometric inconsistency in our ablation study. Besides our 3D point error, we use reprojection error. We also consider the minimum distance between the rays of cameras pointing to the same 3D point, averaged across all 3D points~\cite{schinstock2009alternative,hessflores2011ray}; we refer to this term as to minimal ray distance.
Moreover, we implement BA with inequality constraints, or IBA~\cite{lhuillier2011fusion}. In IBA, we run the first round of BA with only a geometric term that gives a minimum value. We use this value to impose an inequality constraint on the geometric term and run the second round of BA with other loss terms added.

According to Tab.~\ref{tab:ablation_geometric_term}, 3D point error surpasses other geometric terms. Reprojection error provides slightly worse results; this could be expected since it reduces optimization of 3D locations to 2D reprojection space. We also observe that optimizing minimum ray distance leads to a negligible improvement. Surprisingly, the advanced IBA modifications are inferior to the basic terms. We attribute this to numerous keypoint mismatches occurring since the floor and walls are textureless or covered with repetitive patterns. $L_{geom}$ brings matched points closer, so mismatches impose a risk of scan corruption. But if using the floor and walls terms, pulling mismatched points in the wrong directions is compensated by floor-to-plane and walls-to-floorplan pulling enforcing the floor and walls to remain flat.

\begin{figure}[h!]
\centering
\setlength{\tabcolsep}{1pt}
\begin{tabular}{c}
    \includegraphics[width=0.65\linewidth]{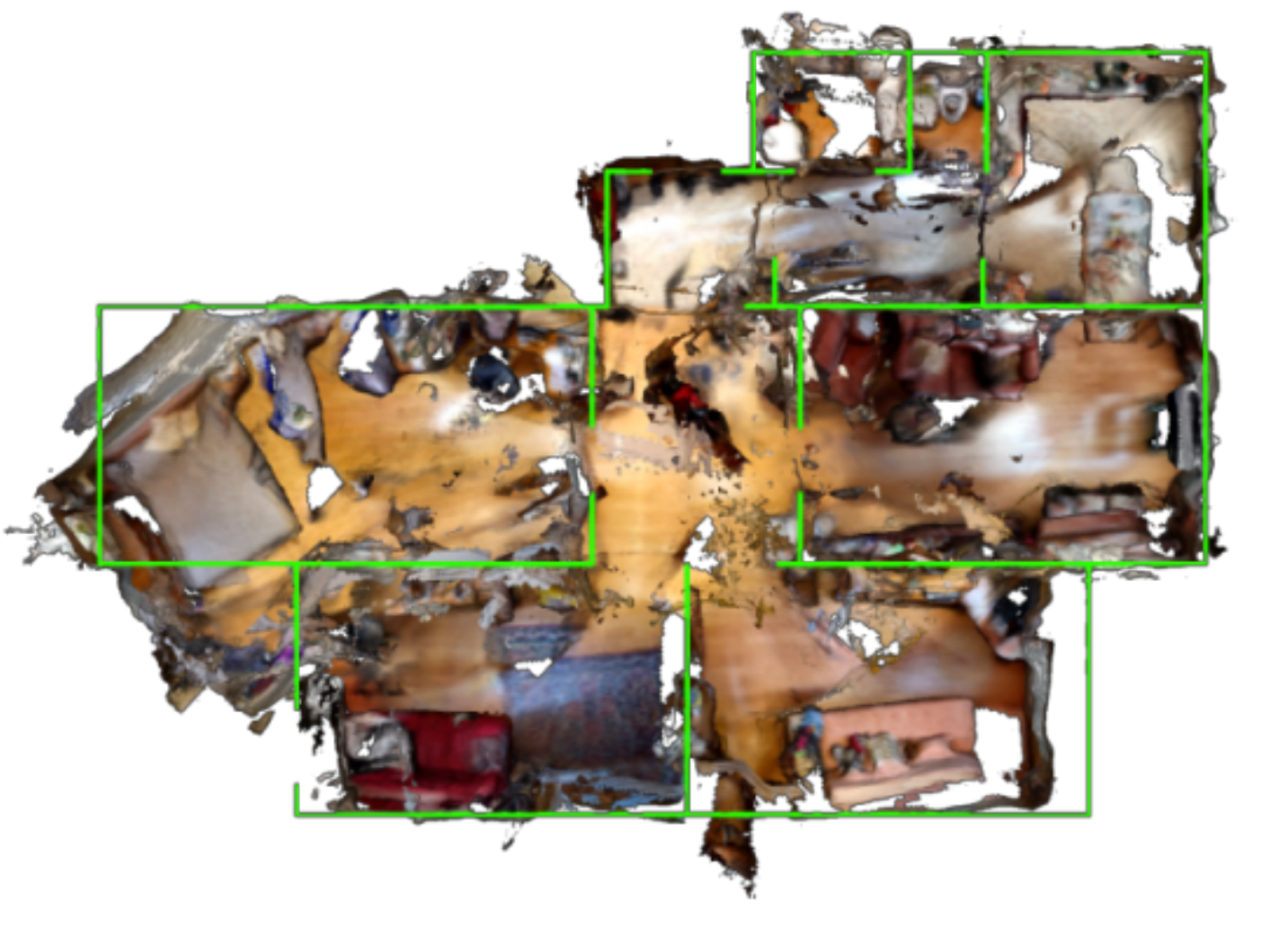} \\
    (a) Nearest point strategy \\
    \includegraphics[width=0.65\linewidth]{images/rechnoy_floor_walls_pulling.pdf} \\
    (b) Fixed nearest plane strategy \\
\end{tabular}
\caption{The nearest point strategy corrupts the scan (a), while the fixed nearest plane strategy allows to recover the general scene structure (b).}
\label{fig:smapper_rechnoy_ablation_walls_term}
\end{figure}

\textbf{Walls term.} We also try different walls terms in combination with geometric (3D point error) and floor terms. As can be seen in Tab.~\ref{tab:ablation_walls_term}, the nearest point strategy allows for the best NSD since it directly minimizes the distance to the floorplan. The iterative nearest plane and the fixed nearest plane strategies aim at more stable pulling and encourage more minor scan corrections in exchange for better reconstruction quality~(Fig.~\ref{fig:smapper_rechnoy_ablation_walls_term}). 

According to the reconstruction metrics (MME, MPV, MOM), the fixed nearest plane strategy notably outperforms other strategies on our self-captured data in reconstruction metrics. This strategy exploits more spatial information, which allows for stable convergence even when camera poses are noisy. In Redwood, camera trajectories are quite precise, so the iterative nearest plane strategy converges while performing on par with the fixed nearest plane strategy. Either way, we claim the fixed nearest plane strategy to be a better option since it is more stable while being faster in the order of magnitude. 

\section{Conclusion}

We proposed a novel camera pose refinement approach expanding the bundle adjustment concept. Our method accepts posed RGB-D frames as inputs and updates camera poses according to the prior scene structure given as a floorplan. Through experiments on Redwood and our own dataset of RGB-D scans, we demonstrated that our approach improves the accuracy of 3D reconstruction from RGB-D data. 






\bibliographystyle{plain}
\bibliography{literature}

\end{document}